\documentclass[a4paper]{article}

\usepackage{INTERSPEECH2019}

\title{Conversational Emotion Analysis via Attention Mechanisms}
\name{Zheng Lian$^{1,3}$, Jianhua Tao$^{1,2,3}$, Bin Liu$^{1}$, Jian Huang$^{1,3}$}
\address{
$^{1}$National Laboratory of Pattern Recognition, CASIA, Beijing, China\\
$^{2}$CAS Center for Excellence in Brain Science and Intelligence Technology, Beijing, China\\
$^{3}$School of Artificial Intelligence, University of Chinese Academy of Sciences, Beijing, China}
\email{\{zheng.lian, jhtao, liubin, jian.huang\}@nlpr.ia.ac.cn}

\begin{document}

\maketitle
\begin{abstract}
Different from the emotion recognition in individual utterances, we propose a multimodal learning framework using relation and dependencies among the utterances for conversational emotion analysis. The attention mechanism is applied to the fusion of the acoustic and lexical features. Then these fusion representations are fed into the self-attention based bi-directional gated recurrent unit (GRU) layer to capture long-term contextual information. To imitate real interaction patterns of different speakers, speaker embeddings are also utilized as additional inputs to distinguish the speaker identities during conversational dialogs. To verify the effectiveness of the proposed method, we conduct experiments on the IEMOCAP database. Experimental results demonstrate that our method shows absolute 2.42\% performance improvement over the state-of-the-art strategies.

\end{abstract}

\section{Introduction}

%Step 1: 什么是contextual emotion ananlysis,
Recently, conversational emotion analysis has attracted attention due to its wide applications in human-computer interaction. Different from emotion recognition in individual utterances, conversational emotion analysis utilizes the relation among utterances to track the user's emotion states during conversations.

%Step 2: three aspects are importat for contextual emotion ananlysis. [multimodalty fusion methods, speaker information and contextual information] ->
To improve the performance of conversational emotion analysis, prior works have been performed mainly in three directions: (1) After extraction of multimodal features, there is a question of how to fuse these features effectively. (2) After multi-modalities fusion, there is another question of how to utilize contextual features to help predict emotion states of the current utterance. (3) The last question is how to model the interaction of different speakers in conversational dialogs.

% Step 3: multimodalty fusion methods, and correxspond appraoches. ->从别人的方法引出你用attention机制计算融合权重的方法。
After extraction of multimodal features, there are mainly three strategies for multi-modalities fusion, namely feature-level fusion, decision-level fusion and model-level fusion \cite{wu2014survey}.
In feature-level fusion, multimodal features are concatenated into a joint feature vector for emotion recognition. Although this method can significantly improve the performance \cite{eyben2012audiovisual,schuller2012avec}, it suffers from the curse of dimensionality. 
Decision-level fusion can eliminate the disadvantage of feature-level fusion. In decision-level fusion, multimodal features are modeled by corresponding classifiers first, and then recognition results from each classifier are fused by weighted sum or additional classifiers \cite{ramirez2011modeling,metallinou2010decision}. However, it ignores interactions and correlations between different modalities. 
To deal with this problem, a vast majority of works are explored toward model-level fusion. It is a compromise for feature-level fusion and decision-level fusion methods, which is proposed to fuse intermediate representations of different features \cite{wu2013two,jiang2011audio,paleari2009evidence}. 
Recently, attention-based model-level fusion strategies have gained promising results. Chen et al. \cite{chen2016multi} proposed a multimodal fusion strategy, which can dynamically pay attention to different modalities at each timestep. Poria et al. \cite{poria2017multi} proposed an attention-based network for multimodal fusion, which fused multimodal features via the attention score for each modality. Inspired by the power of attention mechanisms for emotion recognition \cite{chen2016multi, poria2017multi}, we also utilize the attention-base model-level fusion strategy in this paper.

% Step 4: contetxual infos are important
According to the emotion generation theory \cite{gross2011emotion}, human perceive emotions not only through individual utterances but also by surroundings. To take into account the contextual effect in the dialogs, Vanzo et al. \cite{vanzo2014context} used wider contexts (such as topics and hashtags) to help identify emotion states. Poria et al. \cite{poria2017context} proposed a LSTM-based network to capture contextual information from their surroundings. Recently, the self-attention mechanism \cite{vaswani2017attention} has been verified to attend to longer sequences than many typical RNN-based models \cite{vaswani2017attention,liu2018generating}. This mechanism can provide an opportunity for injecting the global context of the whole sequence into each input utterance. However, this mechanism ignores information about the order of the sequential utterances in conversational dialogs \cite{vaswani2017attention}, which is important for conversational emotion analysis. To deal with this problem, we use the bi-directional gated recurrent unit (GRU) layer \cite{chung2014empirical}, in combination with the self-attention mechanism for emotion analysis. With the help of the bi-directional GRU layer, the order of sequential utterances is injected into hidden vectors. Then the self-attention mechanism is utilized to capture the long-term contextual information.

% Step 5: speaker infomation: ->介绍之前论文speaker infos 很重要，-》 你在论文里面也用到了说话人信息。 (讲一下speaker infos和contextual infos 的区别)
As a person's emotion state can be influenced by the interlocutor's behaviors in conversational dialogs \cite{lee2009modeling}, speaker information can be utilized to improve the performance of emotion recognition. A vast majority of works have been explored to model the interaction of different speakers in recent years \cite{lee2009modeling,zhang2017interaction,chen2017multimodal, zhao2018multi}. Lee et al. \cite{lee2009modeling} proposed a dynamic bayesian network to model the conditional dependency between two interacting partners’ emotion states in a dialog. Zhang et al. \cite{zhang2017interaction} proposed a multi-speaker emotion recognition model. The emotion probabilities of previous utterances of each speaker were utilized to estimate emotion of the current utterance. Chen et al. \cite{chen2017multimodal, zhao2018multi} combined feature sequences of different speakers to imitate the interaction patterns. These methods \cite{lee2009modeling,zhang2017interaction,chen2017multimodal, zhao2018multi} need explicit speaker identity of each utterance. But speaker identities are not always available in sensitive or anonymous conversations. To handle this situation, we propose a new interactive strategy in this paper. We first extract speaker embeddings for each utterance using the pre-trained speaker verification system \cite{snyder2017deep}. Then these embeddings are utilized as additional inputs to distinguish speaker identities in conversations.

\begin{figure*}[h]
	\centering
	\includegraphics[width=0.9\linewidth]{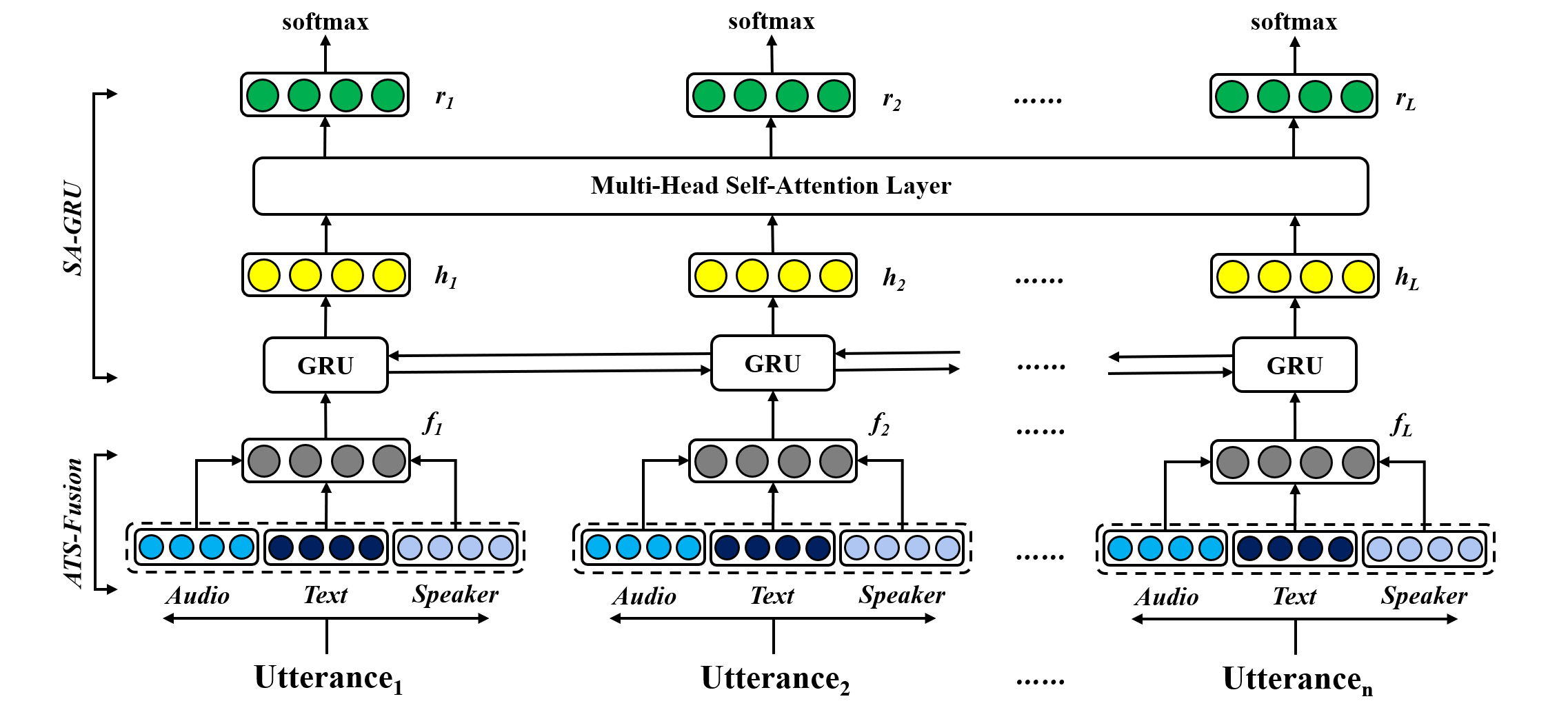}
	\caption{Overall structure of the proposed framework: multimodal input features are fused by ATS-Fusion, followed by SA-GRU for emotion classification.}
	\label{fig:system}
\end{figure*}

% Step 6: 最后，介绍我们的方法， 将这三部分信息融合，对contextual emotion ananlysis 进行建模。同时介绍contribution
This paper proposes a multimodal learning framework for conversational emotion analysis. The main contributions of this paper lie in four aspects: 1) to prioritize important modalities, we use the attention mechanism for multi-modalities fusion; 2) to capture long-term contextual information, we use the bi-directional GRU layer, in combination with the self-attention mechanism; 3) to model the interaction of different speakers when explicit speaker identities are unavailable, we utilize speaker embeddings as additional inputs; 4) our proposed method is superior to state-of-the-art approaches for conversational emotion analysis. To the best of our knowledge, it is the first time that the self-attention mechanism is utilized for conversational emotion analysis. In the meantime, we provide an approach to modeling the interaction when explicit speaker identities are unavailable.

\section{Proposed Method}

To infer the emotion state in realistic conversational dialogs, we propose a multimodal learning framework. As illustrated in Fig. 1, the proposed framework contains two components: the Audio-Text-Speaker Fusion component (ATS-Fusion) for multi-modalities fusion and the Self-Attention based GRU component (SA-GRU) for emotion classification. ATS-Fusion takes the audio, text and speaker information as the inputs and outputs robust representations using the attention mechanism. SA-GRU uses the bi-directional GRU layer, in combination with the multi-head self-attention layer to amplify the important contextual evidents for emotion classification. Meanwhile, to model the interaction of different speakers when explicit speaker identities are unavailable, we utilize speaker embeddings as additional inputs.

\subsection{Speaker encoder}

Speaker embeddings should capture the speaker characteristics. We extract speaker embeddings from each utterance using the pre-trained speaker verification system in \cite{snyder2018xvectors}. This system maps variable-length utterances to fixed-dimensional embeddings, known as x-vector \cite{snyder2018xvectors}. The architecture is based on the end-to-end system described in \cite{snyder2017deep}. It consists of five layers that operate on speech frames, a statistics pooling layer that aggregates over the frame-level representations, two additional layers that operate at the segment-level, and finally a softmax output layer. In the training process, this system also utilizes data augmentation to multiply the amount of training samples and improve robustness.

\subsection{Attention-based multi-modalities fusion: ATS-Fusion}
Not all modalities are equally relevant in emotion classification. To prioritize important modalities, we utilize the attention mechanism \cite{poria2017multi,cho2014learning,luong2015effective} in ATS-Fusion. This method takes audio, text and speaker information as inputs and outputs an attention score for each modality.

Let us assume a video to be considered as $V=[u_{1}, u_{2}, ..., u_{i}, ..., u_{L}]$, where $u_{i}$ is the $i^{th}$ utterance in the video and $L$ is the number of utterances in the video. We first extract acoustic features $a_{i}$, lexical features $t_{i}$ and speaker features $s_{i}$ from the utterance $u_{i}$. Then we equalize the feature dimensions of all three inputs to size $d$ and process utterance-level concatenation:

\begin{equation}
u_{i}^{cat} = Concat(W_{a}a_{i}, W_{t}t_{i}, W_{s}s_{i})
\end{equation}
where $W_{a}$, $W_{t}$ and $W_{s}$ are feature embedding matrices for acoustic features, lexical features and speaker features, respectively. Here, $W_{a}a_{i} \in \mathbb{R}^{d \times 1}$, $ W_{t}t_{i} \in \mathbb{R}^{d \times 1}$, $W_{s}s_{i} \in \mathbb{R}^{d \times 1}$ and $u_{i}^{cat} \in \mathbb{R}^{d \times 3}$.

The attention weight vector $\alpha_{fuse}$ and the fusion representation $f_{i}$ of the input utterance $u_{i}$ are computed as follows:
\begin{equation}
P_{F} = tanh(W_{F}u_{i}^{cat})
\end{equation}
\begin{equation}
\alpha_{fuse} = softmax(w_F^T P_{F})
\end{equation}
\begin{equation}
f_{i} = u_{i}^{cat}\alpha_{fuse}^T
\end{equation}
where $W_{F} \in \mathbb{R}^{d \times d}$ and $w_{F} \in \mathbb{R}^{d \times 1}$ are trainable parameters. Here, $ P_{F} \in \mathbb{R}^{d \times 3}$, $\alpha_{fuse} \in \mathbb{R}^{1 \times 3}$ and $f_{i} \in \mathbb{R}^{d \times 1}$.

\subsection{Self-attention based emotion classifier: SA-GRU}

In the conversational emotion analysis, the emotion state of the target utterance is temporally and contextually dependent to surroundings. To take into account such dependencies, we use the bi-directional GRU layer, in combination with the multi-head self-attention layer for emotion classification.

\subsubsection{Gated recurrent unit}
GRU is usually adopted as the basic unit in RNN as it is able to solve the vanishing gradient problem in the conventional RNN training. Since each GRU cell consists of an update gate and a reset gate to control the flow of information, it is capable of modeling long-term dynamic dependencies.

Let $F=[f_{1}, f_{2},...,f_{t},...,f_{L}]$ be the input to the GRU network, where $f_{t}$ is the fusion representation of the utterance $u_{t}$ (in Sec 2.2) and $L$ is the number of utterances in the video. To obtain contextually dependent utterance representations $H=[h_{1}, h_{2},...,h_{t},...,h_{L}]$, the GRU network computes the hidden vector sequence $H$ from $F$ with the following equations:

\begin{equation}
z_{t} = \sigma_{g}(W_{z}f_{t}+U_{z}h_{t-1}+b_{z})
\end{equation}
\begin{equation}
r_{t} = \sigma_{g}(W_{r}f_{t}+U_{r}h_{t-1}+b_{r})
\end{equation}
\begin{equation}
h_{t} = (1-z_{t}) \circ h_{t-1}+z_{t} \circ \sigma_{h}(W_{h}f_{t}+U_{h}(r_{t}\circ h_{t-1})+b_{h})
\end{equation}
where $\sigma_{g}$ is the sigmoid activation function, $\sigma_{h}$ is the hyperbolic tangent and $\circ$ is element-wise multiplication. $z$ and $r$ are the update gate vectors and reset gate vectors, respectively. $W$, $U$ and $b$ are weight matrices and bias vectors for each gate.

\subsubsection{Multi-head self-attention layer}

To focus on only relevant utterances in emotion classification of the target utterance, a multi-head self-attention layer is used.

\begin{figure}[h]
	\centering
	\includegraphics[width=0.6\linewidth]{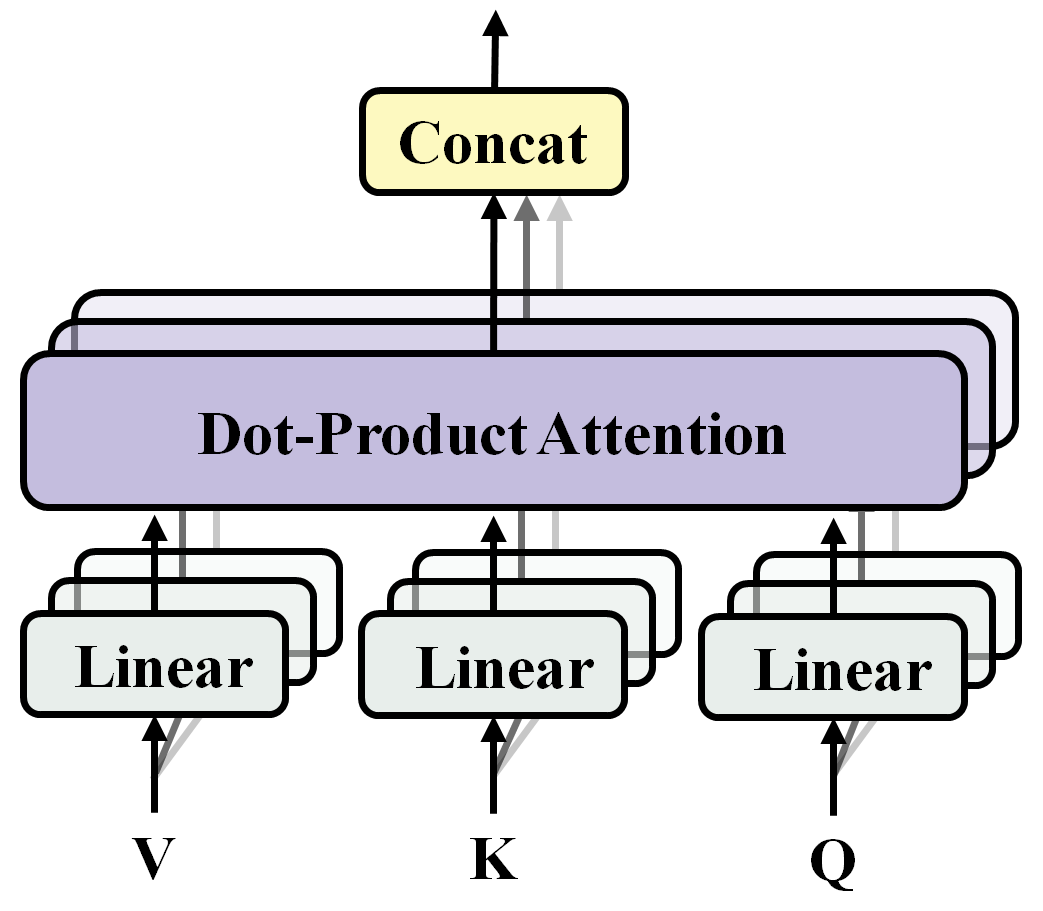}
	\caption{Multi-head self-attention layer: it consists of several dot-product attention layers running in parallel.}
	\label{fig:system}
\end{figure}

The outputs of the GRU network $H=[h_{1}, h_{2},...,h_{L}]$ are passed into the multi-head self-attention layer (in Fig. 2). To extract queries $Q$, keys $K$ and values $V$, we linearly project the output $H$ for $h$ times with different linear projections, which are computed as follows:

\begin{equation}
V = Concat(HW_1^V, ..., HW_h^V)
\end{equation}
\begin{equation}
K = Concat(HW_1^K, ..., HW_h^K)
\end{equation}
\begin{equation}
Q = Concat(HW_1^Q, ..., HW_h^Q)
\end{equation}
where $H \in \mathbb{R}^{L \times d}$, $W_i^Q \in \mathbb{R}^{d \times (d/h)}$, $W_i^K \in \mathbb{R}^{d \times (d/h)}$, $W_i^V \in \mathbb{R}^{d \times (d/h)}$ and $h$ is the number of heads.

On each of these projected versions of queries $HW_i^Q$, keys $HW_i^K$ and values $HW_i^V$, we perform the dot-product attention with following equations:
\begin{equation}
head_i = softmax((HW_i^Q)(HW_i^K)^T)((HW_i^V)
\end{equation}

Then outputs of attention functions $head_i \in \mathbb{R}^{L \times (d/h)}, i \in [1, h]$ are concatenated together as final values $R$. As $R \in \mathbb{R}^{L \times d}$, where $L$ is the number of utterances in a dialog. The matrix $R$ can be represented as $R=[r_{1},r_{2},...,r_{t},...,r_{L}]$, where $r_{t} \in \mathbb{R}^{d}$ for $t=1$ to $L$. Finally, $R$ is fed into a linear projection layer and a softmax layer for emotion recognition.

\section{Experiments and Discussion}

\subsection{Corpus description}
The IEMOCAP \cite{busso2008iemocap} database contains about 12.46 hours of audio-visual conversations in English. There are five sessions with two actors each (one female and one male) and each session has different actors. Each session has been segmented into utterances, which are labeled into ten discrete labels (e.g., happy, sad, angry). To compared with state-of-the-art approaches \cite{poria2017context,rozgic2012ensemble}, we form a four-class emotion classification dataset containing {$happy(1636),angry(1103), sad(1084), neutral(1708)$}, where happy and excited categories are merged into a single happy category. Thus 5531 utterances are involved. To ensure that models are trained and tested on speaker independent sets, utterances from the first 8 speakers are used as the training set and utterances from other speakers are used as the testing set. 

\subsection{Experimental setup}
\textbf{Features}: Acoustic features are extracted from waveforms using the openSMILE \cite{eyben2010opensmile} toolkit. Specifically, we use the IS13-ComParE configuration file in openSMILE. Totally, 6373-dimensional utterance-level acoustic features are extracted, containing Mel-frequency cepstral coefficients (MFCCs), voice intensity, pitch, and their statistics (e.g., mean, root quadratic mean); Lexical features are extracted from transcriptions through two steps. In particular, we first get 1024-dimensional vector representations of words using the deep contextualized word representation — ELMo \cite{peters2018deep}. These word vectors are learned from the deep bidirectional language model trained on the 1 Billion Word Benchmark \cite{chelba2013one}. To obtain utterance-level lexical features, we then calculate mean values of these word vectors; Speaker embeddings are extracted from raw input utterances using the Kaldi Speech Recognition Toolkit \cite{povey2011kaldi}. Specifically, we use the x-vector system \cite{snyder2018xvectors} trained on the NIST SREs \cite{martin2010nist, sadjadi20172016}. Finally, 512-dimensional speaker embeddings are extracted from the x-vector system.

\textbf{Settings}: ATS-Fusion contains three fully-connected layers of size $d=100$. These layers map acoustic, lexical and speaker features into fixed dimensionality, respectively. SA-GRU contains the bi-directional GRU layer with 100 units, followed with a self-attention layer (100 dimensional states and 4 attention heads). We use the Adam optimization scheme with a learning rate of 0.0001 and a batch size of 20. Cross-entropy loss is used as the loss function of the emotion recognition task. To prevent over-fitting, we use dropout \cite{srivastava2014dropout} with $p=0.2$ and $L2$ regularization. To alleviate the impact of the weight initialization, each configuration is tested 20 times. The unweighted accuracy (UA) is chosen as our evaluation criterion.

\subsection {Contribution of individual components}

In this section, we evaluate the contribution of each component in the framework. Four comparison systems with different combination of individual components are implemented to compare with the proposed framework.

$(1)$ System 1 (S1): Ignoring speaker embeddings, we only use acoustic and lexical features for emotion recognition. The attention mechanism in Sec 2.2 is used for multi-modalities fusion, marked as the Audio-Text Fusion component (AT-Fusion).

$(2)$ System 2 (S2): Instead of using ATS-Fusion, we first equalize dimensions of acoustic, lexical and speaker features into $d=100$ via feed forward layers. Then we simply add these multimodal representations together. This fusion approach is marked as $ADD$.

$(3)$ System 3 (S3): Instead of feeding the outputs of ATS-Fusion to SA-GRU, we only use the self-attention layer.

$(4)$ System 4 (S4): Instead of feeding the outputs of ATS-Fusion to SA-GRU, we only use the bi-directional GRU layer.

$(5)$ System 5 (S5): It is our proposed framework for conversational emotion recognition. Acoustic, lexical and speaker features are fused by ATS-Fusion. Then the outputs of ATS-Fusion are fed into SA-GRU for emotion classification.

\begin{table}[h]
	\centering
	\caption{Experimental results with different combination of components.}
	\label{table1}
	\begin{tabular}{ccccc}
		\hline
		& Modalities 	 & Fusion 			 & Classifier 		& UA(\%) 		\\ \hline
		S1 & A+T         & AT-Fusion         & SA-GRU        	& 76.40      	\\
		S2 & A+T+S       & ADD		         & SA-GRU        	& 76.65       	\\
		S3 & A+T+S       & ATS-Fusion        & Attention  		& 66.00       	\\
		S4 & A+T+S       & ATS-Fusion        & Bi-GRU  		    & 77.29       	\\
		S5 & A+T+S       & ATS-Fusion        & SA-GRU        	& 78.02      	\\ \hline
	\end{tabular}
\end{table}

To verify the importance of speaker embeddings, we compare the performance of S1 and S5. Experimental results in Table 1 demonstrate that S5 gains better performance, 78.02\%, than S1, 76.40\%. The speaker embeddings indicate the speaker's identity of each utterance. They reflect personal information and role switches in the conversation, which are important for conversational emotion recognition \cite{zhang2017interaction, chen2017multimodal}. Therefore, cooperating speaker embeddings with other multimodal features can improve the performance of emotion recognition.

To verify the effectiveness of ATS-Fusion, we compare the performance of S2 and S5. Experimental results show that S2 achieves a limited performance in emotion recognition. ATS-Fusion takes audio, text and speaker information as inputs and prioritizes important modalities via attention weights. This approach is similar to human perceptions since humans can dynamically focus on more trustful modalities to understand emotions \cite{chen2016multi, poria2017multi}. While $ADD$ is a special case of ATS-Fusion. It treats each modality equally and cannot select relevant modalities for emotion recognition. Therefore, ATS-Fusion is more suitable for multimodal fusion than $ADD$.

To verify the effectiveness of SA-GRU, we compare the performance of S3, S4 and S5. We find S5 is superior to S3 and S4. The bi-directional GRU layer (in S4) models the temporal and contextual dependence in conversational dialogs. The self-attention layer (in S3) amplifies the important contextual evidences for emotion analysis of the target utterance. To employ the power of these components, our proposed SA-GRU (in S5) combines the bi-directional GRU layer with the self-attention layer for emotion recognition. Experimental results indicate the effectiveness of SA-GRU, which can gain better performance than the single bi-directional GRU layer (or the single self-attention layer).

Meanwhile, experimental results in Table 1 demonstrate that S3 gains much lower performance than S4. The self-attention layer contains no recurrence and no convolution networks. If we shuffle the order of utterances in the dialogs, we will get the same output. However, the order of sequential utterances is important for conversational emotion analysis \cite{vaswani2017attention}. With the help of the GRU layer, the order of sequential utterances is injected into hidden vectors, which can further improve the performance of emotion recognition.

\subsection{Comparison to state-of-the-art approaches}

To show the effectiveness of the proposed method, we compare our method with currently advanced approaches. Experimental results of different methods are shown in Table 2.

\begin{table}[h]
	\centering
	\caption{The performance of state-of-the-art approaches and the proposed approach on the IEMOCAP database.}
	\label{table1}
	\begin{tabular}{lc}
		\hline
		Approaches				 								& UA (\%)	      \\ \hline
		Rozgi{\'c} et al. (2012) \cite{rozgic2012ensemble}   	& 67.40           \\
		Jin et al. (2015) \cite{jin2015speech}   				& 69.20           \\
		Poria et al. (2017) \cite{poria2017context}   			& 75.60           \\
		Li et al. (2018) \cite{li2018inferring}     			& 74.80           \\
		Proposed method							      			& \textbf{78.02}  \\ \hline
	\end{tabular}
\end{table}

Compared with our proposed method, these approaches \cite{poria2017context, rozgic2012ensemble, jin2015speech, li2018inferring} also utilized acoustic and lexical features for emotion recognition. Rozgi{\'c} et al. \cite{rozgic2012ensemble} combined decision trees with support vector machines (SVM) for the sentence-level multimodal emotion recognition. Jin et al. \cite{jin2015speech} investigated different ways to combine acoustic and lexical features, including decision-level fusion and feature-level fusion. Poria et al. \cite{poria2017context} proposed a LSTM based network to capture contextual information from their surroundings in the same video. Li et al. \cite{li2018inferring} proposed a multi-modal, multi-task deep learning framework to infer the user’s emotive states.

Experimental results in Table 2 demonstrate the effectiveness of the proposed method. Our method shows absolute 2.42\% performance improvement over state-of-the-art approaches. It proves that modeling long-term dynamic dependencies and considering the speaker information can improve the performance of emotion recognition.

\section{Conclusions}

This paper proposes a multimodal learning framework to infer the emotion state in realistic conversational dialogs. To evaluate the effectiveness of the proposed method, we conduct experiments on the IEMOCAP database. Experimental results reveal that ATS-Fusion, which is employed to fuse multimodal features, can provide robust representations of each utterance for following SA-GRU. SA-GRU, which amplifies the important contextual evidents for emotion classification, can gain better performance than the single bi-directional GRU layer (or the single self-attention layer). With the help of speaker embeddings, the proposed framework can utilize the interactive information in conversational dialogs, which can further improve the performance of emotion recognition. Experimental results on the IEMOCAP database demonstrate that the proposed framework is superior to state-of-the-art strategies.

\section{Acknowledgements}
This work is supported by the National Key Research \& Development Plan of China (No.2018YFB1005003), the National Natural Science Foundation of China (NSFC) (No.61425017, No.61831022, No.61773379, No.61771472), and the Strategic Priority Research Program of Chinese Academy of Sciences (No.XDC02050100).

\bibliographystyle{IEEEtran}

\bibliography{mybib}

\end{document}